\documentclass[manuscript,screen,twoside,authorversion]{acmart}
\usepackage{graphicx}
\usepackage{listings}
\usepackage{stmaryrd}
\usepackage{xcolor}
\usepackage{graphicx}
\usepackage{float}
\usepackage{tikz}
\usepackage{booktabs}
\usepackage{url}
\usepackage{tabularx}
\usepackage{array}
\usepackage{parskip}

\SetSymbolFont{stmry}{bold}{U}{stmry}{m}{n}
\DeclareFontFamily{U}{stmry}{}
\DeclareFontShape{U}{stmry}{m}{n}{
  <-5.5>  stmary5
  <5.5-6.5> stmary6
  <6.5-7.5> stmary7
  <7.5-8.5> stmary8
  <8.5-9.5> stmary9
  <9.5->  stmary10
}{}

\definecolor{isarblue}{HTML}{006699}
\definecolor{isarlight}{HTML}{0099ff}
\definecolor{isargreen}{HTML}{009966}
\definecolor{isarred}{HTML}{e34013}
\definecolor{lightgray}{gray}{0.95}
\lstdefinelanguage{isabelle}{%
    keywords=[1]{type_synonym,datatype,primrec,fun,abbreviation,definition,proof,lemma,theorem,corollary,value,using,by,proposition, record, locale, inductive_set, unfolding, from, moreover,ultimately,consider, qed, next, then, have},
    keywordstyle=[1]\bfseries\color{isarblue},
    keywords=[2]{fixes,where,assumes,shows,and,obtains,for},
    keywordstyle=[2]\bfseries\color{isargreen},
    keywords=[3]{fix, assume, case, define, show},
    keywordstyle=[3]\color{isarlight},
    commentstyle=\color{isarred},
    morecomment=[s]{(*}{*)},
    moredelim=[is][\color{isarblue}]{|-}{-|}, 
    alsoletter={?},
}
\newcommand*{\circled}[2][]{\tikz[baseline=(C.base)]{
    \node[inner sep=0pt] (C) {\vphantom{1g}#2};
    \node[draw, circle, inner sep=3pt, yshift=1pt]
        at (C.center) {\vphantom{1g}};}}

\begin{document}

\title{IsabeLLM: Automated Theorem Proving Applied to Formally Verifying Consensus}


\author{Elliot Jones}
\affiliation{%
  \institution{Imperial College London}
  \city{London}
  \country{United Kingdom}}
\email{e.jones24@imperial.ac.uk}

\author{William Knottenbelt}
\affiliation{
  \institution{Imperial College London}
  \city{London}
  \country{United Kingdom}}

\begin{abstract}
  Advances in Artificial Intelligence (AI) have led AI for Theorem Proving to become a promising means of formally verifying computer systems. Whilst formal verification is traditionally reserved for safety-critical systems due to the required amount of expertise and effort, AI can help to automate a large amount of this workload and make it far more accessible. Blockchain-based systems are becoming increasingly popular and are frequently targeted by malicious actors, often resulting in huge financial losses, highlighting the need to better verify these systems and mitigate vulnerabilities. Arguably the most important component of these systems is the consensus protocol, which allows nodes to agree on decisions in a potentially adversarial environment. In this paper, we improve upon IsabeLLM, the automated theorem proving tool in Isabelle. Namely, we implement a Retrieval-Augmented Generation framework, Error tracing and counterexample generation for improved context supplied to the Large Language Model. Compatibility with the latest version of Isabelle and Sledgehammer is also implemented for improved efficiency. We compare the performance of the two versions of IsabeLLM in their ability to complete the verification of Bitcoin's Proof of Work consensus.
\end{abstract}

%


\maketitle

\section{Introduction}

A blockchain enables peer-to-peer digital transactions without the need for a trusted intermediary. This is only possible because of its consensus protocol, which allows nodes within the system to agree on the state of the blockchain, even in the presence of adversaries. For this reason, it is paramount that consensus is designed and implemented correctly to prevent the system from reaching unwanted states that can be exploited by adversaries. The most famous example of this is Bitcoin's Proof of Work (PoW) consensus protocol and its susceptibility to a 51\% attack, where adversaries control the majority of the compute power in the system and giving them the potential to double spend. Infamous examples of such attacks include Ethereum Classic~\cite{EthClassic51}, Bitcoin Gold~\cite{BTCgold51}, and Vertcoin~\cite{VertCoin51}, totalling losses of over \$30 million. 

Other key components of modern blockchain systems are bridging protocols for cross-chain data transfer and smart contracts for automated agreement execution. These components are also not without their exploits, with infamous examples such as the Poly Network~\cite{PolyNetworkAttack}, Wormhole Bridge~\cite{WormholeAttack}, Binance Smart Chain~\cite{BNBChainAttack}, and Qubit Finance~\cite{QubitFinAttack}, totalling losses of over \$1.5 Billion. These failures further underscore the need to ensure correctness across the domain.

Formal verification is the process of formalising a system and then mathematically proving its correctness. However, it is often underutilised in the software development process because of the large amount of effort and expertise it requires. Whilst organisations like the Ethereum Foundation actively fund formal verification work in the space, it continues to see the huge financial losses discussed previously. Furthermore, blockchain systems cannot rely on the traditional \lq test and patch\rq model for their consensus protocols as patching would require a hard fork, such as the Ethereum/Ethereum Classic split~\cite{EthHardFork}. Hard Forks are disruptive and controversial as they challenge blockchain's core principle of immutability. Furthermore, patching smart contracts is often impossible once they have been deployed on a blockchain as they are usually immutable~\cite{SmartContractsImmutable} with some exceptions~\cite{SmartContractsUpgradeable}. This outlines the need for formal verification, as it can be used for correctness-by-construction~\cite{CorrectByConstruct} and minimise the need for costly post-deployment fixes.

In recent years, the field of Artificial Intelligence has made incredible progress, particularly within the realm of Large Language Models (LLMs) like OpenAI's ChatGPT and High-Flyer's DeepSeek. This advancement has opened up new opportunities across all domains, including the formal verification space. In particular, AI for theorem proving has gained traction and has started to see applications outside of purely mathematical statements and instead for program verification. An example of this is FVEL~\cite{aitpFVEL}, which is used to assist in automated verification of C/C++ programs in the Isabelle proof assistant. This reduces the entry barrier for the formal verification of such programs, making it more accessible and less time-consuming.

In our original paper, we presented IsabeLLM~\cite{IsabeLLM}, a tool that integrates the proof assistant Isabelle with a Large Language Model to assist and automate proofs. We demonstrated the effectiveness of IsabeLLM by using it to prove the correctness of a novel model for Bitcoin's Proof-of-Work consensus protocol.

We expand on the original paper by introducing IsabeLLM-RAG, an improved version of IsabeLLM. Specifically, we made the following improvements:

\begin{enumerate}
    \item Implemented a Retrieval-Augmented Generation (RAG) database of proofs from the binary tree model in~\cite{ElliotDiego} for improved domain-specific context.
    \item Implemented the Nitpick Counterexample Generator~\cite{blanchette2010nitpick} to identify logically false proof steps.
    \item Implemented an error trace that captures all modifications between LLM calls, providing the history of changes as context and highlighting which proof steps have been corrected and should be preserved.
    \item Compatibility with Isabelle 2025 and general efficiency improvements.
    \item Changed LLM from DeepSeek R1 to DeepSeek R1T2 Chimera for improved efficiency.
\end{enumerate}

We discuss these changes in more detail in Section~\ref{sec:improv}.

\section{Background}

\subsection{Blockchain}

A blockchain is a decentralised ledger that allows multiple parties to carry out transactions without the need of a trusted intermediary, eliminating the need for trust. This is only possible through a blockchain's consensus protocol, which allows all parties to agree on the current state of the blockchain and the transactions recorded on it. The most popular consensus protocol is Proof of Work (PoW) used by Bitcoin's blockchain, which has around 1.2 billion recorded transactions~\cite{BtcTransactions} with Bitcoin's market capitalisation sitting around \$1.8 trillion~\cite{BtcMarketCap}. The core idea of PoW is that the longest blockchain is correct since it assumes the majority of computing power within the system is honest and therefore should be able to solve hashes and add blocks faster than adversaries~\cite{BtcWhitepaper}.

\subsection{Isabelle}

Isabelle is a proof assistant written in Scala and ML that uses Higher-Order Logic (HOL). It is used to write and verify formal proofs with high assurance due to the mechanisation of these proofs~\cite{Isabelle}. Isabelle's Isar proof language allows these proofs to be more readable than the traditional approach to theorem proving by repeatedly applying tactics. Isabelle also makes use of automation tools like Sledgehammer, which uses external automated theorem provers (ATPs) to help you complete proofs. Outside of the proof assistant itself, the Scala library Scala-Isabelle provides the functionality to interact with an Isabelle process inside of a Scala application~\cite{scalaIsabelle}.

\subsection{Retrieval-Augmented Generation}

Retrieval-Augmented Generation (RAG) is the technique of using external data to improve the performance of LLMs. Although LLMs are great generalist tools, they often struggle in niche domains where there is little available data, resulting in the LLM hallucinating an incorrect answer. RAG aims to combat this by providing a database of domain-specific data to help give the LLM more context and understand the problem.

A RAG framework consists of a retrieval mechanism and a generative LLM. Data is first mapped into a vector space such that the closer data points are to one another, the more similar they are. The retriever searches this vector space to find the most relevant information to a given query and then includes it in the prompt fed to the LLM~\cite{RAGSeminal}.

In the context of AI for Theorem Proving, RAG is used to identify similar statements to what we are currently trying to prove. By seeing how similar statements have been proven in the past, it helps the LLM understand how to prove the new statement. RAG-based techniques have been used in both Lean~\cite{thakur2023copra} and Isabelle~\cite{first2023baldur} for purely mathematical proofs but has yet to be used in the context of formal verification.

\section{Related Work}

Isabelle has been used extensively in the last 20 years to carry out numerous verifications. Some of the most notable verifications include the seL4 Microkernel~\cite{isabelleSel4}, the ML compiler~\cite{isabelleCakeML}, and numerous protocol and program verifications~\cite{isabelleDistributedSystems,isabelleHybridSystems,isabelleSAT}. Outside of verification, Isabelle has been used to formalise a large amount of mathematics that can be found in the Archive of Formal Proofs (AFP)~\cite{isabelleAFP}. Some of the most notable formalisms in the AFP include G{\"o}del's incompleteness theorems~\cite{isabelleGodel}, Jordan curve theorem~\cite{isabelleJordanCurve}, and Ramsey's Theorem~\cite{isabelleRamsey}. In recent years, Isabelle has been used for verifications and formalisms of blockchain systems, including the Ethereum Virtual Machine~\cite{isabelleEVM} and a framework to verify solidity smart contracts~\cite{isabelleSolidity}.

Outside of Isabelle, various other theorem provers have been used to carry out verifications in the blockchain domain. To name a few, Agda~\cite{agdaBtc,agdaBtcScript,agdaSolidity}, Coq~\cite{coqSmartContract,coqEthSmartContract,coqAlgorand}, and Lean~\cite{leanClear,LeanAMM} have been used for the formalisms of blockchain. The field has also started to see formalisms of Decentralised Finance (DeFi) components~\cite{FormalDeFi,FormalAMM,FormalEV} but are yet to be mechanised. Other major works within the space include a formalism of the Ethereum Virtual Machine in the K~\cite{kevm}, Certora Prover~\cite{certoraProver} for semi-automated smart contract verification, and fuzzer Echidna~\cite{echidna}. Most recently, PropertyGPT~\cite{propertygpt} used LLMs trained on Certora auditing reports to formally verify smart contracts. Zero-Knowledge Proof circuits, which are widely used in blockchain systems, have also seen increased interest in formal analysis~\cite{zkproofAnalysis}.

The field of AI for theorem proving has seen the development of major data sets for proof assistants in recent years, including IsarStep and PISA for Isabelle~\cite{aitpIsarStep,aitpPISA}, LeanDojo for Lean~\cite{aitpLeanDojo}, and GamePad and CoqGym for Coq~\cite{aitpCoqGym}. Using these datasets has allowed for the development of various theorem proving models, including LEGO-Prover~\cite{aitpLegoProver}, LISA~\cite{aitpPISA} and DeepSeek-Prover~\cite{aitpDeepSeekProver}. Artificial Intelligence for formal verification has seen limited use, with the aforementioned FVEL~\cite{aitpFVEL} being the major work in this area. As for AI for formal verification of blockchain, the literature is sparse and has only seen research into extracting smart contract specifications from natural language~\cite{aitpSmartContractSpecs}. 
\section{Model}
\label{sec:mod}

Our consensus model builds on previous work~\cite{ElliotDiego} by generalising the blockchain structure from a binary tree to an n-ary tree. This extension enables the model to account for an arbitrary number of forks at any given block, reflecting a more realistic view of a blockchain. We prove that consensus holds in a majority honest network using the common prefix and chain quality properties outlined in the Bitcoin Backbone Protocol~\cite{btcBackbone}, where they are discussed in more detail. We make the same assumptions of majority honesty and synchronisation in the network, meaning that the majority of the computing power in the network is honest and that everyone shares the same view of the blockchain. Moreover, we assume block difficulty remains constant, simplifying Bitcoin's heaviest chain rule to the longest chain rule. We make these assumptions to reduce complexity in our model and prevent the need to switch to a computational model. Whilst such models offer a more realistic verification, the objective of this research is to assess the efficacy of LLMs in verification. This model serves as a sufficient baseline for future verifications of increased complexity.

As in the previous work, we can omit the chain quality property under our majority honesty assumption, leaving us with the common prefix property which states that all honest parties agree on a common chain up to the last {$k$} blocks in a chain. This is a safety property, showing honest nodes do not diverge except near the tip of the chain. We first define some preliminaries before specifying our consensus theorem.

\begin{definition}[$n$-ary Trees and Height Function]
Let $\mathcal{T}$ be the set of $n$-ary trees. A tree $t \in \mathcal{T}$ is defined as a tuple $t = (x, ts)$ where $x$ is the node ID and $ts = [t_1, t_2, \dots, t_k]$ is a finite list of subtrees, such that each $t_i \in \mathcal{T}$ and $k \ge 0$. Let $Height : \mathcal{T} \to \mathbb{N}$ be a function returning the height of a tree.
\end{definition}

\begin{definition}[Depth Check Function]
For depths $n, d \in \mathbb{N}$, the function $Check : \mathbb{N} \times \mathbb{N} \times \mathcal{T} \to \mathbb{B}$ is defined as follows:
\[ Check(0, d, t) = True \]
\
\[ Check(n+1, d, (x, ts)) = \exists p \in ts : \left( Height(p) > d + \max_{p' \in ts \setminus \{p\}} Height(p') \land Check(n, d, p) \right) \]

This function ensures that no other branch is less than $d$ nodes away from the longest chain up to depth $n$ of the tree.
\end{definition}

\begin{definition}[Events]
An event $e \in \mathcal{E}$ is defined as a tuple $(b,t) \in \mathbb{B} \times \mathcal{T}$, where $b$ is a boolean flag indicating if the event was honest, and $t$ is the state of the tree immediately after the event. $\mathcal{L}(\mathcal{E})$ denotes the set of all finite sequences of events, representing the traces of the blockchain.
\end{definition}

\begin{definition}[Traces]
Let $add_h : \mathcal{T} \to \mathcal{T}$ and $add_d : \mathcal{T} \to \mathcal{T}$ be the transition functions for honest (adding to the longest chain) and dishonest (adding to any branch) block additions, respectively. Let $State: \mathcal{E} \to \mathcal{T}$ extract the tree from an event, $hd: \mathcal{L}(\mathcal{E}) \to \mathcal{E}$ return the most recent event in a trace, and $C_{bool} : \mathcal{L}(\mathcal{E}) \to \mathbb{N}$ return the number of events in a trace matching the specified boolean flag.

Given a starting tree $t_0$ under the assumption that $Check(n, Height(t_0)-n+1, t_0)$ holds, the inductive set of valid blockchain traces $\mathcal{S} \subseteq \mathcal{L}(\mathcal{E})$ is defined by the following rules:
\begin{enumerate}
    \item \textbf{Honest Base Case:}
    \[ [(True, add_h(t_0))] \in \mathcal{S} \]

    \item \textbf{Dishonest Base Case:}
    \[ [(False, add_d(t_0))] \in \mathcal{S} \]

    \item \textbf{Honest Inductive Step:} For any existing trace $tr \in \mathcal{S}$:
    \[ (True, add_h(State(hd(tr)))) :: tr \in \mathcal{S} \]

    \item \textbf{Dishonest Inductive Step:} For any existing trace $tr \in \mathcal{S}$, given the condition $C_{F}(tr) < C_{T}(tr) + (Height(t_0) - n)$ holds:
    \[ (False, add_d(State(hd(tr)))) :: tr \in \mathcal{S} \]
\end{enumerate}
\end{definition}

\begin{theorem}[Consensus]
\label{thm:consensus}
Let $LongestPaths: \mathcal{L}(\mathcal{E}) \to \mathcal{P}(\mathcal{T})$ return the set of the longest branches, and let the function $take: \mathbb{N} \times \mathcal{T} \to \mathbb{N}$ return the prefix of a branch up to the specified depth $n$. For all valid traces, all longest paths share the same prefix up to depth $n$:
\[ \forall tr \in \mathcal{S}, \forall p, p' \in LongestPaths(State(hd(tr))) : take(n, p) = take(n, p') \]
\end{theorem}


This statement is identical to the consensus statement for the binary tree model. However, the generalisation to an n-ary tree significantly increases the complexity of the proof. In the binary tree case, inductive arguments typically require only two cases (e.g., left and right subtrees), whereas the n-ary setting necessitates reasoning over an arbitrary number of branches, complicating case distinctions and inductive reasoning. To show this, Table~\ref{tb1} gives a description for each lemma and the Lines of Proof (LoP) required to complete the verification of each model. We only list the lemmas that were more than one LoP in at least one of the models. In the binary tree column, ``N/A'' means that the lemma was not required for the verification. For the rows with \(x+y\), \(x\) is the LoP that are `original' and \(y\) is the LoP that are symmetric to $x$ and are just repeated for the different cases. With this in mind, it is clear that the n-ary tree model has more than double the original LoP when compared to the binary tree model.


\begin{table}[htbp]
\centering
\begin{tabular}{@{} l l l p{7.5cm} @{}}
\toprule
\textbf{Lemma Name} & \textbf{Binary Tree} & \textbf{N-ary Tree} & \textbf{Description} \\
\midrule
subtree\_height & N/A & 15 & The height of all trees in a set are less than or equal to the highest tree in the set. \\
\midrule
height\_mono & 1+1 & 23 & Monotonicity of tree heights. \\
\midrule
obtain\_max & N/A & 23 & In a set of trees, there exists a tree that is at least as high as all other trees. \\
\midrule
foldr\_max\_eq & N/A & 37 & The maximum height of a set of trees is equal to the height of the tallest tree. \\
\midrule
branch\_height & N/A & 30 & The height of a branch of a tree is less than or equal to the height of the tree. \\
\midrule
sub\_longest & N/A & 28 & Given the longest branch in a tree, then there exists a subtree where its prefix is a longest branch. \\
\midrule
sub\_branch & N/A & 41 & Given the longest branch in a tree, then its prefix is a longest branch for the prefix of the tree. \\
\midrule
weaken\_distance & 1 & 18 & For any tree $t\in\mathcal{T}, $ $Check(n,d+1,t), \to Check(n,d,t)$ \\
\midrule
weaken\_depth & 1 & 15 & For any tree $t\in\mathcal{T}, $ $Check(n+1,d,t), \to Check(n,d,t)$  \\
\midrule
common\_prefix & 25+12 & 38 & If two branches are longest branches in a tree and no other node exists at a distance less than $d$ with a common ancestor on the path to the root, then the two branches share a common prefix up to depth $n$. \\
\midrule
height\_add\_mining & 10+5 & 36 & When mining on a tree, either the maximum height of the tree increases (added to the longest branch) or it stays the same (added to other branch). \\
\midrule
check\_add\_mining & 49+158 & 1 & Given $Check(n,d+1,t)$, then there are 3 cases when mining on the tree:\newline
- Add to longest chain  ($Height(t') +1 = Height(t)$) \newline
- Add to next longest chain ($Height(t) - Height(t') = d+1$)\newline
- Add to any other chain ($Height(t) - Height(t') > d+1$) \\
\midrule
height\_add\_honest & 10+5 & 32 & Mining honestly on a tree always results in the tree height increasing (mining on longest chain). \\
\midrule
check\_add\_honest & 22+13 & 36 & For tree $t\in\mathcal{T}, $ $Check(n,d,t), \to Check(n,d+1,add_h(t))$  \\
\midrule
bounded\_check & 56 & 17 & For any trace $tr \in \mathcal{L}(\mathcal{E})$ and $t_0 \in \mathcal{T}$, \newline
$Check(n,C_{T}(tr) + Height(t_0)-n - C_{F}(tr),State(hd(tr)))$\\
\midrule
consensus & 1 & 5 & As stated in Theorem~\ref{thm:consensus}. \\
\midrule
\textbf{Total} & \textbf{175+193} & \textbf{395} & \\
\bottomrule
\end{tabular}
\vspace{10pt}
\caption{Lines of Proof (LoP) and Lemma Descriptions for each tree model.}
\label{tb1}
\end{table}

\newpage
\mbox{}
\newpage

\begin{figure}[htbp]
  \centering
  \includegraphics[width=0.8\linewidth]{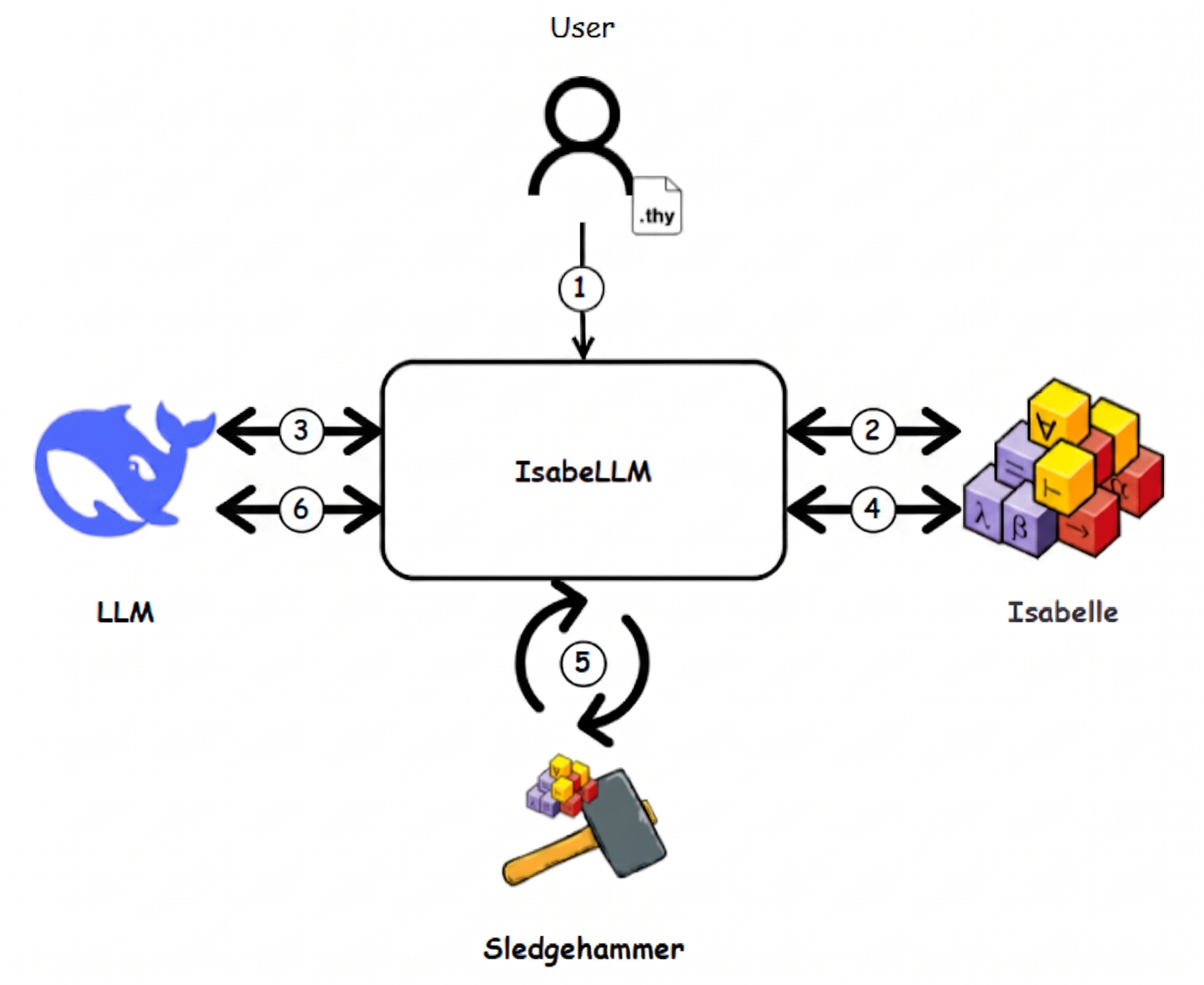}
  \caption{IsabeLLM Architecture.}
  \label{fig:arch}
  \Description{Shows the interface between Isabelle, Sledgehammer and the LLM.}
\end{figure}

\section{IsabeLLM}

IsabeLLM is an interface between the Isabelle proof assistant and an LLM. It is designed for general purpose and so can be used to prove any kind of statements within Isabelle. It should be noted that if you are using bespoke imports for your theory file, then they should be given to the LLM as context for it to understand. In our models, we are only importing Isabelle's Main library, and everything is contained within the single theory file, meaning we do not need to provide extra context.

\begin{table} [htbp]
\centering
\begin{tabularx}{\textwidth}{|>{\hspace{0pt}}l<{\hspace{0pt}}|>{\hspace{0pt}}X<{\hspace{0pt}}|}
\hline
\textbf{Error} & \textbf{Description} \\
\hline
``Sorry'' Detected in Proof & 
The sorry keyword is used to mark incomplete proofs. IsabeLLM uses sorry to detect on which part of the theory file to work on.
\\
\hline
Failed Proof &
The proof failed. IsabeLLM attempts to Sledgehammer before retrying the LLM.
\\
\hline
Undefined Fact/Method &
LLM hallucinated a fact or method. Hallucinations are removed before rebuilding.
\\
\hline
Lexical/Syntax Error &
Bad syntax injected into the theory file. IsabeLLM has various methods to detect these issues but will return the proof back to the LLM if it fails.
\\
\hline
Timeout &
Build exceeds time limit, often due to hanging tactics (e.g., metis, blast). IsabeLLM verifies these tactics using Sledgehammer.
\\
\hline
\end{tabularx}
\vspace{10pt}
\caption{Isabelle build errors.}
\label{tb2}
\end{table}

\subsection{Architecture}
\label{sec:arch}
 
The high-level architecture for IsabeLLM can be seen in Fig.~\ref{fig:arch}. The main idea is that we use an LLM to understand the high-level structure of a proof and then use Isabelle's Sledgehammer tool to solve the intermediate steps that the LLM failed (if any). The general workflow for IsabeLLM is as follows:

\begin{enumerate}
    \item [\circled{1}] The user uploads their Isabelle theory file (.thy) to their working directory, along with a ROOT file so that the Isabelle server knows which files to look at. The user starts IsabeLLM.
    \item[\circled{2}] IsabeLLM first uses the Isabelle server to try and build the theory file. If there are no issues with the file and all statements have been proven, then the build completes, and we are done. If not, then IsabeLLM captures the errors raised to identify the unproven statements and extracts them.
    \item[\circled{3}] IsabeLLM sends the context of the theory file and the unproven lemma to the LLM via its API. The LLM tries to prove the lemma and returns a proof of the statement.
    \item[\circled{4}] IsabeLLM injects the new proof into the theory file and tries to build it again. If this fails, we send the file to Isabelle's Sledgehammer tool.
    \item[\circled{5}] Sledgehammer tries to solve each unproven line within the proof. If some are left unproven, then IsabeLLM extracts these lines and their errors, along with the rest of the updated theory file.
    \item[\circled{6}] IsabeLLM returns the current proof state to the LLM and asks it to resolve the remaining errors. The LLM returns a proof of the statement.
    \item [\circled{7}] Steps 4--6 are repeated until the theory file is successfully built or IsabeLLM reaches a set number of iterations.
\end{enumerate}

\begin{figure}[htbp]
  \centering
  \includegraphics[width=0.7\linewidth]{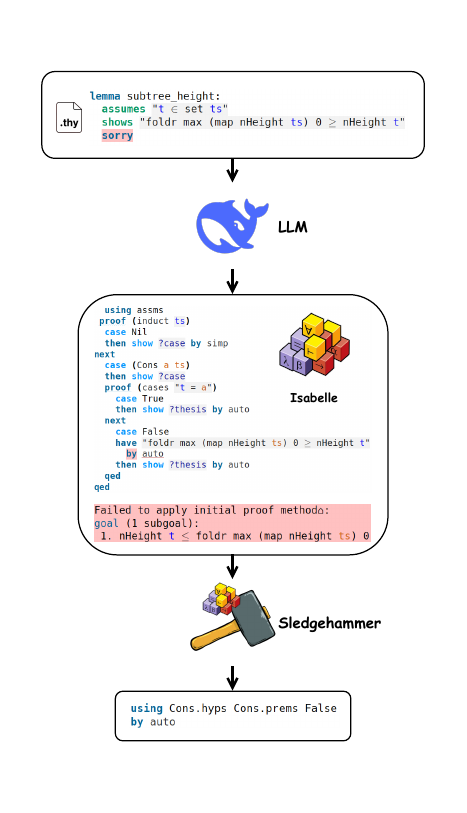}
  \caption{IsabeLLM example workflow.}
  \label{fig:example}
  \Description{We prove the subtree_height lemma.}
\end{figure}

Fig.~\ref{fig:example} shows an example workflow in IsabeLLM. In this example, we prove the lemma subtree\_height which states that the height of a tree in a set of trees is always greater than or equal to the maximum height of the set of trees. The LLM generates a proof that fails a proof step, which we then correct with Sledgehammer. The proof is by induction over the list of trees and splitting the inductive step into the cases of whether the tree is at the head of the list or not.

For this paper, we opted to use variations of the DeepSeek R1 model as our LLM due to its strong coding benchmarks and free access to its API. Claude Sonnet was also considered, but was ultimately decided to be too expensive. Other models like OpenAI's GPT-4 and Mistral's Le Chat were also considered but showed poor performance during manual testing. As for our choice of proof assistant, we chose Isabelle due to its existing automation tool Sledgehammer and integration library Scala-Isabelle. Although not integral, the Isar language also helps to understand the logic of the proofs and the dialogue between IsabeLLM and LLM.

\subsection{Implementation}
\label{sec:imp}

Almost all of IsabeLLM is written in Scala, with some Python to access the LLM API using the OpenAI library. The main reason for choosing Scala is to be able to use the Scala-Isabelle library, which offers the functionality to control an Isabelle process from a Scala application. In particular, we make use of Scala-Isabelle for calling Sledgehammer in our theory file. All of our code, including the IsabeLLM source code and theory files, can be found at~\cite{isabellmRepo}. We list IsabeLLM's features below:

\begin{enumerate}
    \item Interface between Isabelle and a LLM API.
    \item Code extraction from a theory file, including lemmas, definitions, and proofs.
    \item Injection of code into a theory file, including lemmas, definitions, and proofs.
    \item Sledgehammer functionality with a timeout control and option to select which provers it uses. In this paper, we set this timeout to 60 seconds and use the default provers.
    \item Handling of errors in the build process. An Error trace is maintained between LLM calls for improved context.
    \item Records and updates to the LLM chat history.
    \item LLM Prompt Generation.
    \item RAG Database for improved context sent to LLM.
    \item Nitpick counterexample generator to identify logically false statements.
    \item Controls for the maximum number of LLM iterations before timing out to prevent endless loops. 
    
\end{enumerate}

Due to the stochastic nature of LLMs, the most challenging part of automating proof with IsabeLLM is ensuring the output from the LLM has the correct syntax to be injected into Isabelle. Generally speaking, most models understand the syntax well enough to give a coherent response. However, most outputs will trigger at least one error in the build process and must be handled accordingly. Table~\ref{tb2} highlights the general types of error that are encountered when trying to build the theory file after injecting a generated proof.

When sending requests to the LLM, IsabeLLM automatically builds the required prompts to make the context clear. When we first initialise a proof, we send a prompt that includes the context of the lemma we are trying to prove and everything in the theory file before it. After the initialisation prompt, we send prompts that include only the current proof state of the lemma and error trace. Similar proofs extracted from the RAG database are appended to the end of both messages. The templates for these prompts can be seen in Table~\ref{prompts}. IsabeLLM also maintains the chat history with the LLM. To minimise the size of our context, we reset the history after successfully proving a lemma, then update the initial context to the theory file with the updated lemma.

\begin{table}[htbp]
\centering
\begin{tabularx}{\textwidth}{|>{\hspace{6pt}}l<{\hspace{6pt}}|>{\hspace{6pt}}X<{\hspace{6pt}}|}
\hline
\textbf{Prompt} & \textbf{Text} \\
\hline
Initialisation & 
I am trying to complete a proof in Isabelle. Here is my theory file so far:  (.thy file). I am trying to prove the following lemma: (lemma). Please prove this lemma. Return only the raw code without any additional text, explanations, formatting, or commentary. Do not include ``` or language tags. Just the pure code.
\\
\hline
Error &
Your proof is incorrect. The current proof state is: (proof state). The line: (error line) produced the following error message: (error). Please amend the proof to deal with this error. Since your last output, the system attempted the following automated fixes which resulted in the current state: (error trace). Return only the raw code without any additional text, explanations, formatting, or commentary. Do not include ``` or language tags. Just the pure code.
\\
\hline

\end{tabularx}
\vspace{10pt}
\caption{IsabeLLM prompts.}
\label{prompts}
\end{table}

\subsection{Improvements}
\label{sec:improv}

This paper improves on the original IsabeLLM tool by making a variety of changes to improve performance and efficiency, most notably by enabling stronger context in our prompts. The first of these is the implementation of a Retrieval-Augmented Generation (RAG) framework. To ensure the LLM receives self-contained context, we process the Isabelle .thy files using regular expressions to isolate definitions and lemmas. We then build a dependency graph and perform a breadth-first search to identify all transitive dependencies (e.g., datatypes and functions) that a lemma relies on. These dependencies are recursively bundled with the proof text before vectorisation. Embeddings are generated using the SentenceTransformers all-MiniLM-L6-v2 model~\cite{sentencetransformer} and stored in a local ChromaDB instance. The database consists of all 22 proofs (trivial and non-trivial) from the previous binary tree model in~\cite{ElliotDiego}. We query the vector database to find the three most similar lemmas using Euclidean distance as our metric. These lemmas are then injected into the LLM prompt under a dedicated reference header.

\begin{figure}[htbp]
  \centering
  \includegraphics[width=0.7\linewidth]{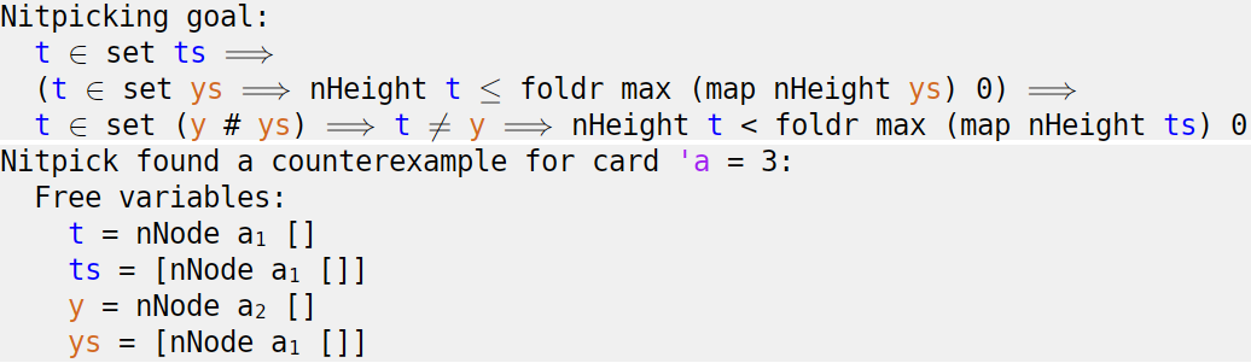}
  \caption{Nitpick Example}
  \label{fig:nitpick}
  \Description{}
\end{figure}

We also integrated the Nitpick counterexample generator~\cite{blanchette2010nitpick}. In the previous paper we found that LLMs would repeatedly try to prove statements that were logically false. Nitpick identifies these statements which we can then provide as context to the LLM to help it understand why the proof does not work and adapt to a new approach. Fig.~\ref{fig:nitpick} shows an example counter example. Here, the LLM is trying to prove the intermediate step that the height of tree $t$ is strictly less than the maximum height of all trees in the set $ts$. However, nitpick identifies a counterexample where $t$ has the maximum height and so the statement must be false.

Lastly, we implemented an error trace that is added context to the LLM. In the previous paper we noted the LLM would often override changes we made between LLM calls and get stuck in a loop by repeatedly changing proof steps we have already corrected. The trace provides the exact changes that have been made since the last LLM call and highlights what should be preserved, enabling the LLM to build on our improvements instead of trying to start from scratch each call.

Next, we aimed to improve the efficiency of IsabeLLM. First, we made the tool compatible with Isabelle 2025. The original tool used Isabelle 2022 as this was the version that the Scala-Isabelle package~\cite{scalaIsabelle} supported. Since then, the package has been updated to be compatible with the latest Isabelle versions and its associated tools, including Sledgehammer. In the previous paper we noted that the older version of Sledgehammer failed to find proofs in the allotted time but the Isabelle 2025 sledgehammer succeeded. Updating to the latest version enables us to use these more efficient tools, improving our ability to find proofs. Moreover, the original IsabeLLM ran different processes for the Isabelle Server and Sledgehammer. We have updated this to consolidate the whole tool into a single process, massively improving the speed of computation and significantly reducing the required memory.

\subsection{Model Selection}

In this paper, we test IsabeLLM-RAG with 3 different models. Namely, these are DeepSeek R1T2 Chimera~\cite{r1t2chimera},  NVIDIA-Nemotron-3-Super-120B-A12B~\cite{nemotron} and OpenAI's gpt-oss-120b~\cite{gptoss}. We summarise their architectures in Table~\ref{tab:model_architecture}. The original IsabeLLM paper used DeepSeek R1 as its model, using a standard Mixture-of-Experts (MoE) transformer architecture with 671B total parameters and around 37B active parameters, optimised via reinforcement learning for pure reasoning. In contrast, DeepSeek R1T2 Chimera maintains the same parameter footprint but employs an Assembly of Experts (AoE) methodology~\cite{AoE2025}, combining the R1, V3-0324, and R1-0528 models to stabilise outputs. It also offers a slightly larger context window. NVIDIA Nemotron 3 Super introduces a MoE Hybrid Mamba-Transformer architecture with 120B total and 12B active parameters, leveraging LatentMoE routing~\cite{latentmoe} and Multi-Token Prediction (MTP)~\cite{multitokenpred} to achieve a 1M token context window. Finally, GPT-OSS-120B uses a traditional MoE transformer with 117B total and 5.1B active parameters, with MXFP4 quantisation and a 128K context window.

Our DeepSeek models are by far the largest in terms of total parameters, roughly 5 times the size of our NVIDIA and OpenAI models. From this, we would expect our DeepSeek models to outperform the others. Nemotron and GPT-OSS are of similar size, with Nemotron being the most recent model and making use of recent innovations to enable a massive context window. Moreover, GPT-OSS is quantised with MXFP4, meaning it operates at lower precision to significantly reduce its memory footprint, while activating far fewer parameters (5.1B) than our other models.

While the Chimera model is arguably less advanced compared to the pure reasoning of the R1 model, it is much more stable and efficient. The reason for this switch is that we noted one of the main efficiency bottlenecks of the tool is the slow response time of the LLM. In this case, the speed increase is worth the slight decrease in performance, especially when our other improvements help mitigate this decrease.

While the baseline R1 model is highly optimised for pure mathematical reasoning, we hypothesised that the AoE methodology in the Chimera architecture would provide greater syntax stability and inference efficiency. As noted, a primary bottleneck of IsabeLLM is the latency of LLM generation. Reviewing our OpenRouter API logs, DeepSeek R1's speed typically ranged between $10-20$ tokens per second (tok/s). In contrast, Chimera's speed typically ranged between $30-50$ tok/s, representing a significant latency improvement. Furthermore, as demonstrated in Section~\ref{sec:res}, Chimera also minimised the number of corrective loops required, achieving the lowest average number of LLM iterations per successful proof (1.06 compared to 1.31 for R1). Both Nemotron and GPT-OSS exhibited a similar generation speed to the Chimera model.

\begin{table}[htbp]
\centering
\caption{Model Architectures}
\label{tab:model_architecture}
\renewcommand{\arraystretch}{1.3}
\resizebox{\textwidth}{!}{%
\begin{tabular}{@{}llccc@{}}
\toprule
\textbf{Model} & \textbf{Architecture} & \textbf{Total Parameters} & \textbf{Active Parameters} & \textbf{Context Window} \\ 
\midrule
\textbf{DeepSeek R1} & MoE Transformer & 671B & $\sim$37B & 128K \\
\textbf{DeepSeek R1T2 Chimera} & AoE Transformer & 671B & $\sim$37B & 163.8K \\
\textbf{NVIDIA Nemotron 3 Super} & MoE Hybrid Mamba-Transformer & 120B & 12B & 1M \\
\textbf{GPT-OSS-120B} & MoE Transformer & 117B & 5.1B & 128K \\ 
\bottomrule
\end{tabular}%
}
\end{table}

\section{Results}
\label{sec:res}

We tested IsabeLLM and IsabeLLM-RAG by using them to prove each of the 16 non-trivial lemmas listed in Table~\ref{tb1}. We attempt each proof 10 times with a maximum of 3 iterations per attempt, not counting instances when the LLM would return an empty output as this is an API failure rather than a methodological failure. We reduced the maximum number of iterations from 5 in the first paper as we discovered that the additional iterations were almost always ineffective. We consider an attempt to be a failure if it exceeds the maximum number of iterations or exits prematurely (often due to syntax issues). It should be noted that the LoP specified for each lemma can vary as the LLM can generate different proofs for the same statement.

The original IsabeLLM used DeepSeek R1 as its LLM. To evaluate the impact of underlying model architecture on our retrieval-augmented framework, IsabeLLM-RAG was tested using DeepSeek R1T2 Chimera, NVIDIA-Nemotron-3-Super-120B-A12B, and OpenAI's gpt-oss-120b. As discussed in Section~\ref{sec:improv}, IsabeLLM-RAG benefits from the RAG database, error trace, counterexample generator, and access to the Isabelle 2025 Sledgehammer. All experiments were conducted on a workstation running Windows 11. The system was equipped with an AMD Ryzen 7 5825U processor (8 cores, 16 threads) and 16.0 GB of RAM.

Table~\ref{tb3} shows the results of each version of IsabeLLM. Recall that in the previous paper, for the lemma branch\_height, we had to manually use the 2025 Sledgehammer to succeed. These manual interventions have been removed so that we can more clearly see the automated improvements in IsabeLLM-RAG.

Generally speaking, IsabeLLM-RAG outperforms the original across all lemmas when using the Chimera and Nemotron models. It matches or improves the success rate for nearly every lemma while decreasing the average number of iterations per successful attempt. Overall, the Chimera model achieved a 94.4\% success rate with a significant reduction in LLM iterations. Nemotron achieved a 87.5\% success rate, outperforming IsabeLLM despite it using a much larger model. The most notable improvement is for the lemma branch\_height, which previously failed without manual Sledgehammer intervention. IsabeLLM-RAG (Chimera) achieved a 90\% success rate here, needing only 1 LLM call in almost every successful attempt. Conversely, the GPT-OSS model struggled, yielding a 67.5\% average success rate and frequently requiring the maximum number of iterations. However, this is not far behind how IsabeLLM performed with DeepSeek R1, a much larger model without the precision loss of quantisation. 

Similar to the original paper, IsabeLLM-RAG rarely obtained the correct answer on the very first LLM generation for complex lemmas; IsabeLLM had to apply various error corrections and leverage Sledgehammer to fully complete the proofs. Interestingly, where the original paper saw some variation in proof approaches, particularly for larger proofs, IsabeLLM-RAG produced highly deterministic structural approaches across iterations.

\begin{table}[htbp]
\centering
\caption{Performance comparison between IsabeLLM Versions.(SR = Success Rate, Iter. = Avg. Iterations)}
\label{tb3}
\renewcommand{\arraystretch}{1.1} 
\setlength{\tabcolsep}{4pt} 

\resizebox{\textwidth}{!}{%
\begin{tabular}{@{} l c cc cc cc cc @{}}
\toprule
 & & \multicolumn{2}{c}{\textbf{IsabeLLM}} & \multicolumn{6}{c}{\textbf{IsabeLLM-RAG}} \\
\cmidrule(lr){3-4} \cmidrule(l){5-10}
 & & \multicolumn{2}{c}{\textbf{R1}} & \multicolumn{2}{c}{\textbf{Chimera}} & \multicolumn{2}{c}{\textbf{Nemotron}} & \multicolumn{2}{c}{\textbf{GPT-OSS}} \\
\cmidrule(lr){3-4} \cmidrule(lr){5-6} \cmidrule(lr){7-8} \cmidrule(l){9-10}
\textbf{Lemma Name} & \textbf{Lines} & \textbf{SR} & \textbf{Iter.} & \textbf{SR} & \textbf{Iter.} & \textbf{SR} & \textbf{Iter.} & \textbf{SR} & \textbf{Iter.} \\
\midrule
subtree\_height      & 15 & 100\% & 1.0 & 100\% & 1.0 & 100\% & 1.0 & 100\% & 1.0 \\
height\_mono         & 23 & 100\% & 1.0 & 100\% & 1.0 & 100\% & 1.0 & 100\% & 1.0 \\
obtain\_max          & 23 & 90\%  & 1.4 & 100\% & 1.0 & 100\% & 1.0 & 100\% & 1.0 \\
foldr\_max\_eq       & 37 & 50\%  & 2.0 & 100\% & 1.0 & 80\%  & 1.1 & 30\%  & 3.0 \\
branch\_height       & 30 & 0\%   & N/A & 90\%  & 1.1 & 50\%  & 1.2 & 60\%  & 2.0 \\
sub\_longest         & 28 & 70\%  & 1.1 & 90\%  & 1.1 & 100\% & 1.0 & 40\%  & 1.0 \\
sub\_branch          & 41 & 50\%  & 1.8 & 80\%  & 1.1 & 50\%  & 1.4 & 40\%  & 1.0 \\
weaken\_distance     & 18 & 100\% & 1.0 & 100\% & 1.0 & 100\% & 1.0 & 100\% & 1.0 \\
weaken\_depth        & 15 & 100\% & 1.0 & 100\% & 1.0 & 100\% & 1.0 & 100\% & 1.0 \\
common\_prefix       & 38 & 60\%  & 1.5 & 70\%  & 1.4 & 50\%  & 1.2 & 0\%   & N/A \\
height\_add\_mining & 36 & 60\%  & 2.0 & 80\%  & 1.1 & 70\%  & 2.0 & 20\%  & 2.0 \\
check\_add\_mining & 1  & 100\% & 1.0 & 100\% & 1.0 & 100\% & 1.0 & 100\% & 1.0 \\
height\_add\_honest & 32 & 80\%  & 1.7 & 100\% & 1.2 & 100\% & 1.5 & 100\% & 2.5 \\
check\_add\_honest  & 36 & 90\%  & 1.2 & 100\% & 1.0 & 100\% & 1.0 & 40\%  & 2.0 \\
bounded\_check       & 17 & 70\%  & 1.0 & 100\% & 1.0 & 100\% & 1.0 & 50\%  & 1.1 \\
consensus            & 5  & 100\% & 1.0 & 100\% & 1.0 & 100\% & 1.0 & 100\% & 1.0 \\
\midrule
\textbf{Average}     & \textbf{-} & \textbf{76.3\%} & \textbf{1.31} & \textbf{94.4\%} & \textbf{1.06} & \textbf{87.5\%} & \textbf{1.15} & \textbf{67.5\%} & \textbf{1.44} \\
\bottomrule
\end{tabular}%
}
\end{table}

\section{Discussion}

We first examine the variance in success rates across the models. The five lemmas that IsabeLLM-RAG (Chimera) failed to prove with a 100\% success rate have the largest LoP, showing that increased logical complexity inherently reduces success rate. Exceptions to this are the honest mining variants of height\_add and check\_add, which both require a high number of LoP yet achieved perfect completions across Chimera and Nemotron. This success is likely attributable to the RAG system. As discussed in Section~\ref{sec:mod}, the two tree models have different approaches to proof; however, the logic of adapting the proofs from the mining variant to the honest mining variant remains consistent in both models. The LLM identifies this structural symmetry from the RAG context and applies it to prove these lemmas reliably. On the other hand, three of the failed proofs were unique to the n-ary tree model, meaning there was no analogous proof in the binary tree database. It is evident that the RAG context is less effective here as it cannot provide a good template.

As expected, our largest model Chimera performed best, achieving the highest success rate and lowest average iterations. Interestingly, Nemotron performed exceedingly well despite its small size. Despite having a large context window, our prompt lengths rarely exceeded 12K tokens, indicating its success does not stem from context capacity. We attribute its performance primarily to its usage of MTP, which forces the model to learn the broader structural syntax of Isar proofs rather than prioritising myopic next-token prediction. GPT-OSS's weaker performance highlights its architectural limitations. Its lower active parameter count and quantisation appear to restrict its capacity for deep reasoning, context retention and syntax consistency.

The lower variability observed in LLM responses is attributable to RAG. The use of RAG effectively prunes the potential proof space, providing a strict template on how to approach the proof and drastically reducing the chance of hallucination. Despite the original DeepSeek R1 having an advantage in pure mathematical reasoning, we saw no instance where it outperformed the Chimera model. However, this is also influenced by other improvements in IsabeLLM-RAG.

In terms of overall efficiency, IsabeLLM-RAG exceeded its predecessor. One of the highlighted issues in the original paper was the LLM getting stuck in a proof loop, repeatedly failing on the same step. Proof loops were effectively mitigated across the Chimera and Nemotron tests due to the stronger context supplied by Nitpick and the error trace, allowing the LLM to learn from immediate failure. It was also noted that the API interactions were far more reliable.

As before, IsabeLLM is still limited to proof completion, unable to generate the initial properties to be proven. This means that the user must specify a statement before it can be verified, including setting up functions, locales, and sets. However, this issue will be largely mitigated if IsabeLLM is used in conjunction with existing verification frameworks rather than building from the ground up. For example, Isabelle/Solidity~\cite{isabelleSolidity} builds most of the model automatically, leaving only the invariant properties to be specified. A remaining challenge is ensuring the LLM understands the bespoke calculus of such frameworks, given the scarcity of relevant proof corpora.

While the primary objective of IsabeLLM-RAG is the full automation of the verification process, the system inherently facilitates proof discovery. All proofs use Isabelle's Isar proof language that is highly readable, allowing users to follow the logic of the proof and gain insight into potentially new approaches. Furthermore, the integration of Nitpick enables the discovery of previously unconsidered edge cases through counterexample generation, effectively highlighting logical flaws in proof steps.

While our current RAG implementation provides static templates based on prior successful proofs, scaling IsabeLLM to larger verifications involving multiple Isabelle theory files proves to be a challenge. A large context becomes infeasible to include in our prompt due to the limited context window, causing our RAG methodology to break down. In this scenario, integrating the Model Context Protocol (MCP)~\cite{anthropic_mcp} presents a complementary solution to our RAG strategy. Whereas RAG effectively anchors the overarching architecture of a proof by matching structural templates, a MCP-based approach would act as a dynamic retrieval mechanism. By equipping the LLM with MCP tools, the model could actively query Isabelle’s environment during inference. This dynamic retrieval would mitigate the context window bottleneck by fetching only the necessary facts.

\section{Conclusion}

In this paper, we improved the original IsabeLLM tool with IsabeLLM-RAG, implementing a RAG database, counterexample generator (Nitpick), improved conversational error trace, and compatibility with Isabelle 2025 and its associated tools. We evaluated IsabeLLM-RAG's performance using three models - DeepSeek R1T2 Chimera, NVIDIA Nemotron 3 Super, and GPT-OSS-120B. With these improvements, we were able to outperform the original tool across the board using the Chimera and Nemotron architectures, increasing success rates and decreasing the average number of LLM iterations required. Although GPT-OSS showed weaker performance, it was not far behind the performance of the  original IsabeLLM tool, which used a much larger model.

While our results highlight IsabeLLM's capacity to evolve alongside state-of-the-art AI and theorem-proving infrastructure, several avenues for future work remain. IsabeLLM could be parallelised to query different LLMs simultaneously, generating a distributed proof tree to maximise the probability of finding a valid proof path. Furthermore, the LLMs could be fine-tuned on domain-specific data, such as the Archive of Formal Proofs or bespoke consensus verifications. Moreover, IsabeLLM could be integrated with semantic embeddings of programming languages within Isabelle, the most prominent in the blockchain space being Isabelle/Solidity. This would transition IsabeLLM from verifying abstract state-machine models to verifying concrete smart contract bytecode, solidifying its viability in real-world, safety-critical applications. Conversely, because IsabeLLM remains a general-purpose proof assistant wrapper, its methodologies could be readily translated to formal verifications in domains entirely outside the blockchain space.

\bibliographystyle{ACM-Reference-Format}
\bibliography{Main}

\appendix

\section*{Appendix}
All relevant code for IsabeLLM can be found at:

\begin{center}
\url{https://github.com/EllbellCode/IsabeLLM}
\end{center}

The repository includes:
\begin{itemize}
  \item Source code for IsabeLLM.
  \item Isabelle theory files for the n-ary tree PoW model.
  \item Setup instructions.
\end{itemize}


\end{document}